\documentclass[11pt]{article}
\usepackage[margin=1in]{geometry}
\usepackage[T1]{fontenc}
\usepackage{lmodern}
\usepackage{microtype}
\usepackage{amsmath,amssymb,amsthm}
\usepackage{booktabs}
\usepackage{array}
\usepackage[skip=6pt,font=small]{caption}
\usepackage{graphicx}
\usepackage{xcolor}
\usepackage[numbers,sort&compress]{natbib}
\usepackage[colorlinks=true,linkcolor=blue!50!black,citecolor=blue!50!black,urlcolor=blue!50!black]{hyperref}
\usepackage{enumitem}
\usepackage{tikz}
\usetikzlibrary{positioning,arrows.meta,fit,backgrounds,calc}
\setlist{nosep,leftmargin=*}

\newtheorem{proposition}{Proposition}
\newcommand{\sys}{\textsc{Lightfoot}}

\title{\sys: Graph Retrieval-Augmented Generation with\\ Exact, Model-Free Document Deletion}

\author{Uday Allu \\ Yellow.ai}
\date{}

\begin{document}
\maketitle

\begin{abstract}
Graph-based retrieval-augmented generation (GraphRAG) improves multi-hop question answering, but deployed systems pay costs that research comparisons rarely measure: model calls at query time, the cost of building the graph, and the cost and correctness of removing documents.
We present \sys, a GraphRAG system in which every entity, link, fact and synonym edge is a deterministic function of per-passage extractions that carry chunk-level provenance.
Retrieval makes no language-model call, indexing makes one per passage and builds the cheapest index of the systems we compare, 26--31\% below HippoRAG~2, and deleting a document is a database transaction with no model calls: 14 seconds for 10\% of an 11{,}402-passage corpus, after which the graph is identical to a fresh index of the surviving documents. LightRAG needed 2{,}187 model calls and 3.7 hours for the same deletion, and fast-graphrag cannot delete.
The conditions for exactness are simple, and a fingerprint audit checks them: HippoRAG~2's deletion is also exact on the benchmark corpora, but on a synthetic corpus of near-duplicate names, modelled on account or product catalogues, its capped neighbour lists leave 565 fewer edges than a rebuild. The same corpus showed that exact deletion does not imply correct entities: \sys's threshold links chained distinct account names into shared entities until we added a pairwise number guard, which leaves retrieval unchanged within 0.1 points.
Under identical extraction, embedding and answer models on held-out questions of MuSiQue, 2WikiMultiHopQA and HotpotQA, \sys{} is competitive with HippoRAG~2 in retrieval: with Holm-corrected paired tests it is better on MuSiQue at recall@5 and recall@10 (+4.8 and +5.5 points) and on 2WikiMultiHopQA at recall@10 (+3.4), with no significant difference elsewhere. Most of the 2WikiMultiHopQA gain comes from seeding the walk with entities named in the question. With the same cross-encoder, dense retrieval trails \sys{} by 9.3 and 17.8 points recall@5 on the two multi-hop sets, which indicates that the gain comes mainly from the graph.
Answers given the same prompt tie on all three datasets, so the practical case for \sys{} rests on query cost, index cost and deletion rather than accuracy.
Code and per-question results: \url{https://github.com/udayallu-ml/Lightfoot-GraphRAG}.
\end{abstract}

\section{Introduction}

Retrieval-augmented generation \citep{lewis2020rag} grounds a language model in documents retrieved at query time.
Plain similarity search often fails on multi-hop questions whose evidence is spread over passages that the question never names.
GraphRAG systems address this by extracting entities and relations into a graph and traversing it at query time \citep{edge2024graphrag,gutierrez2024hipporag,gutierrez2025hipporag2,guo2024lightrag}; surveys catalogue dozens of variants \citep{han2025graphragsurvey,zhang2025graphragsurvey}.

Published comparisons focus on answer accuracy. Three properties that decide whether a system can be operated in an organisation receive little attention.
The first is the \emph{query path}: 16 of the 17 open-source GraphRAG frameworks we reviewed call a language model during retrieval in their default configuration, to extract keywords or entities from the question or to filter candidate facts, which adds latency, cost and a failure point to every query.
The second is \emph{deletion}. Retention policies, contract expiry and the right to erasure \citep{gdpr2016} can require removing a document and everything derived from it.
Summaries that blend information from many documents make this expensive or impossible: a merged entity description must be regenerated, and a community report cannot be split back into the contributions of its sources.
The third is \emph{index cost}, which differs by an order of magnitude across frameworks because of second extraction passes, model-written entity summaries and community reports.

We reviewed the source code of 17 open-source GraphRAG frameworks (Appendix~\ref{app:survey}). Five offer document-level deletion, and none of these documents or tests that the result equals a graph that never contained the document.

\paragraph{Contributions.}
\begin{enumerate}
  \item A design discipline for GraphRAG in which every derived record is a pure function of the surviving per-passage extractions (\S\ref{sec:erasure}): no merged model-written descriptions, no model judge, and no top-$k$ neighbour lists. Deletion is then exact by construction and needs no model calls. We verify this on one corpus at full scale (11{,}402 passages) and on a synthetic corpus, apply the same fingerprint audit to HippoRAG~2, and show a corpus of near-duplicate names on which its deletion diverges from a rebuild. The same corpus shows that exactness does not imply correct entities, and we add a pairwise guard that keeps both.
  \item A retrieval pipeline with no model calls, Personalized PageRank seeded from similar facts and from entities named in the question and fused with a local cross-encoder (\S\ref{sec:retrieval}), which is competitive with HippoRAG~2 under identical models on held-out questions: better on MuSiQue and at the deeper ranks of 2WikiMultiHopQA, tied on HotpotQA. Reranker-controlled baselines indicate that the gain on multi-hop data comes mainly from the graph walk rather than the cross-encoder.
  \item A deletion-audited comparison against three open-source baselines under identical models, with an identical answer stage, reranker controls, held-out tuning and multiple-comparison correction (\S\ref{sec:results}).
  \item Negative and diagnostic results: a model-based entity-merge judge, a hybrid keyword and embedding search channel, and larger answer contexts gave no reliable gain, and two silent baseline failure modes surfaced during benchmarking (\S\ref{sec:analysis}).
\end{enumerate}

\section{Related work}

\paragraph{Community-summary GraphRAG.}
Microsoft GraphRAG \citep{edge2024graphrag} clusters an extracted entity graph and asks a model to write a report for each community. This supports corpus-level questions but makes indexing the most expensive in our survey and ties every report to many source documents.

\paragraph{Keyword dual-level retrieval.}
LightRAG \citep{guo2024lightrag} extracts entities and relations with model-written descriptions, merges duplicate descriptions with further model calls, and retrieves by matching model-extracted keywords against entity and relation vectors.

\paragraph{PageRank over extracted facts.}
HippoRAG \citep{gutierrez2024hipporag} and HippoRAG~2 \citep{gutierrez2025hipporag2} seed Personalized PageRank \citep{haveliwala2002topic} from entities linked to the query and rank passages by the resulting mass. HippoRAG~2 matches the query against fact embeddings and filters candidate facts with a model call. fast-graphrag \citep{circlemind2024fastgraphrag} also ranks with PageRank, after one structured extraction call per chunk.

\paragraph{Other designs.}
RAPTOR \citep{sarthi2024raptor} builds a tree of recursive summaries. KAG \citep{liang2024kag} plans retrieval over a schema-constrained graph with logical forms. \sys{} is closest to the PageRank family but differs in three ways: it makes no model call during retrieval, it fuses a cross-encoder with the walk, and its derived state is provenance-exact.

\paragraph{Incremental graph maintenance.}
Most work on keeping a knowledge graph current addresses growth.
LightRAG merges the graph of a new document into the existing one \citep{guo2024lightrag}; EraRAG inserts new chunks locally into a hashed multi-layer graph so that updates touch a small region \citep{zhang2025erarag}; iText2KG resolves and integrates entities as documents arrive \citep{lairgi2024itext2kg}; and Zep's temporal graph marks facts as superseded rather than removing them \citep{rasmussen2025zep}.
In each of these, the result of an update depends on what was indexed before and in what order, so removing a source is either unsupported or approximate.
\sys{} instead makes every derived record a function of the live per-chunk extractions only, which makes deletion the inverse of insertion.

\paragraph{Deletion and unlearning.}
Deletion-efficient learning \citep{ginart2019forget} and sharded retraining \citep{bourtoule2021sisa} remove training data from models; neural unranking does the same for learned rankers by gradient updates \citep{hou2024unranking}.
For language models, the right to be forgotten has been analysed as a legal and technical problem \citep{zhang2023rtbf}, and one line of work implements forgetting by editing the retrieval corpus rather than the model \citep{wang2025unlearnrag}.
At the vector layer, streaming approximate-nearest-neighbour indexes support deletes with approximate repair \citep{singh2021freshdiskann}, and HNSW deletes are known to leave the index measurably different from a rebuild \citep{xiao2024hnsw}.
These methods are approximate or concern model weights and vectors. \sys{} targets the derived graph layer, where the model-written state of other GraphRAG systems lives, and requires its result to equal a rebuild exactly, which we test by fingerprinting.

\section{Method}
\label{sec:method}
\begin{figure}[t]
\centering
\resizebox{\linewidth}{!}{%
\begin{tikzpicture}[
  node distance=4mm and 5mm,
  box/.style={draw, rounded corners=2pt, align=center, font=\footnotesize, minimum height=11mm, text width=26mm, fill=blue!6},
  llm/.style={box, fill=orange!25, draw=orange!70!black, thick},
  data/.style={box, fill=gray!12, sharp corners},
  arr/.style={-{Latex[length=2mm]}, semithick},
  lane/.style={font=\small\bfseries, anchor=east}]
\node[data] (doc) {Passage};
\node[llm, right=of doc] (ext) {Extract\\(1 LLM call):\\entities, facts,\\evidence};
\node[box, right=of ext] (emb) {Embed names,\\facts, passage};
\node[box, right=of emb] (link) {Link forms:\\cosine $\ge 0.92$\\+ number guard};
\node[box, right=of link] (comp) {Entities =\\connected\\components};
\node[box, right=of comp] (syn) {Synonym edges:\\cosine $\ge 0.80$,\\no top-$k$ cap};
\node[data, right=of syn] (store) {Store + graph\\(provenance on\\every record)};
\draw[arr] (doc)--(ext); \draw[arr] (ext)--(emb); \draw[arr] (emb)--(link); \draw[arr] (link)--(comp); \draw[arr] (comp)--(syn); \draw[arr] (syn)--(store);
\node[data, below=20mm of doc] (q) {Question};
\node[box, right=of q] (qe) {Embed question\\(bge-base)};
\node[box, right=of qe, text width=46mm] (seed) {Seeds: entities of top-2 facts;\\entities named in the question;\\100 dense passages ($\times 0.05$)};
\node[box, right=of seed] (ppr) {Personalized\\PageRank,\\damping 0.9};
\node[box, right=of ppr] (rr) {Top 30 passages:\\cross-encoder,\\RRF ($k{=}60$)};
\node[llm, right=of rr] (ans) {Answer\\(1 LLM call)\\from top 10 passages};
\draw[arr] (q)--(qe); \draw[arr] (qe)--(seed); \draw[arr] (seed)--(ppr); \draw[arr] (ppr)--(rr); \draw[arr] (rr)--(ans);
\draw[arr, dashed] (store.south) -- ($(store.south)+(0,-6mm)$) -| node[pos=0.25, above, font=\scriptsize] {graph, facts, alias table} (seed.north);
\node[data, below=12mm of q] (del) {Delete\\documents};
\node[box, right=of del] (rm) {Remove their\\chunks, forms,\\fact sources};
\node[box, right=of rm, text width=36mm] (rec) {Recompute touched\\components and synonym edges};
\node[box, right=of rec] (cache) {Purge caches\\(extractions,\\vectors, traces)};
\node[data, right=of cache, text width=40mm] (eq) {Store = fresh index of the survivors (fingerprint audit)};
\draw[arr] (del)--(rm); \draw[arr] (rm)--(rec); \draw[arr] (rec)--(cache); \draw[arr] (cache)--(eq);
\node[lane, left=3mm of doc] {Index};
\node[lane, left=3mm of q] {Query};
\node[lane, left=3mm of del] {Delete};
\end{tikzpicture}}
\caption{The \sys{} pipeline. Orange steps call a language model; all other steps are deterministic and model-free. Every stored record depends only on the live per-passage extractions, so deletion recomputes only what the deleted passages touched and leaves the same store as re-indexing the survivors (Proposition~1).}
\label{fig:pipeline}
\end{figure}
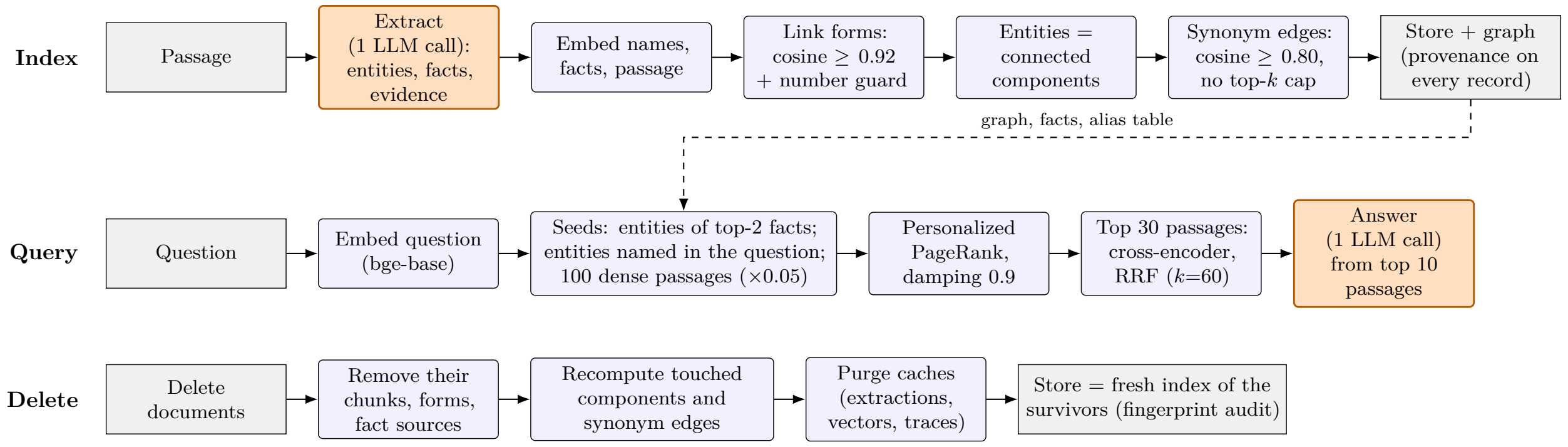

Figure~\ref{fig:pipeline} shows the three paths through \sys: indexing, querying and deletion. Only two steps call a language model, extraction at index time and the final answer; everything between them, including all of retrieval and deletion, is deterministic and model-free.

\subsection{Extraction}
Each passage is processed by exactly one call to an open 70B instruction model \citep{grattafiori2024llama3}. The call returns minified JSON with up to 12 entities, each with a name, a type from a configurable list and a description of at most 8 words. It also returns up to 12 facts, each written as an array of subject, predicate, object and a verbatim evidence quote of at most 12 words.
Writing facts as arrays and forbidding pretty-printing reduced output from 436 to 269 tokens per passage on a 32-passage sample. Over the same sample, facts per passage rose from 3.8 to 7.7.
On MuSiQue the extractor averaged 442 input and 310 output tokens per passage.
Extractions are cached under a key that includes a hash of the prompt and schema.

\subsection{Provenance model and the erasure invariant}
\label{sec:erasure}
Each occurrence of a name in a chunk is stored as a \emph{form} $(c, n, t)$: chunk $c$, normalised name $n$, type $t$, plus that chunk's description of the entity.
Two forms of the same type are \emph{linked} when the cosine similarity of their name embeddings is at least $\tau_m = 0.92$, unless both names contain digits and their sorted digit sequences differ (the \emph{number guard}: ``14 June'' and ``15 July'', or ``Customer account 0001'' and ``Customer account 0005'', are never linked).
Name embeddings are computed in fixed-shape batches, so a name's vector does not depend on which other names share its batch.
An \emph{entity} is a connected component of live forms. Its identifier is derived from its lexicographically smallest member name, and its canonical name is the alias with the most provenance, with ties broken alphabetically.
\emph{Synonym edges} join every pair of same-type entities whose name similarity is at least $\tau_s = 0.80$, without a top-$k$ cap.
\emph{Facts} are keyed by their subject and object forms and carry the set of chunks that asserted them.

\begin{proposition}
Let $G(D)$ denote the store produced by indexing document set $D$. For any partition $D = A \cup B$, deleting $B$ from $G(D)$ yields a store equal to $G(A)$.
\end{proposition}
\begin{proof}[Proof sketch]
Every stored record is a function of the set of live forms and their chunks.
Forms and fact assertions are per-chunk and are removed with their chunk.
Form links, including the number guard, depend only on the two forms involved. Components therefore depend only on the set of live forms, never on insertion order, batching or history, and so do entity identifiers and canonical names.
Synonym edges are a threshold function of the live entity names, with no top-$k$ lists that could depend on removed neighbours.
No record is produced by a model call that reads several chunks.
Deleting $B$ removes exactly the forms and assertions contributed by $B$'s chunks, then recomputes the components and synonym edges it touched. That recomputation yields the same records as computing them from $A$ alone.
\end{proof}

The proposition is nearly a tautology once its premise holds; the substance lies in making the premise hold, which common GraphRAG components violate.
A model-written entity or relation description that merges several chunks, as in LightRAG and Microsoft GraphRAG, is a function of every chunk that contributed to it and of the order of merges.
A model judge for ambiguous name pairs makes a link depend on what the judge saw.
A top-$k$ nearest-neighbour synonym list, as in HippoRAG~2, makes an entity's edges depend on neighbours that may later be deleted: once they are gone, the next-nearest survivors are not promoted.
\sys{} therefore drops all three. Descriptions stay per chunk, the judge was removed after it changed no retrieval metric beyond noise (\S\ref{sec:analysis}), and synonym edges use a pure threshold.
Section~\ref{sec:results} shows that this discipline is what separates \sys's deletion from HippoRAG~2's, whose deletion is also free of model calls.
None of these violations is hard to repair once found: HippoRAG~2's could be fixed by recomputing the neighbour lists of every entity whose list contained a deleted entity. Our claim is not an algorithm that others lack, but a checklist that makes exactness hold by construction, and a fingerprint audit that detects the violations; our own earlier version failed that audit (\S\ref{sec:results}).
Exact erasure says nothing about whether the entities are right. Single-linkage components can chain distinct names into one entity, and a guard against that must itself be a function of the two names alone, as the number guard is.

\paragraph{Deletion in practice.}
A deletion runs as one transaction over all requested documents, so each affected entity component and its synonym edges are recomputed once. Because every record depends only on the live forms, this batched pass leaves the same store as deleting the documents one at a time; a unit test checks both against a fresh index.
For erasure requests, data derived from a document must also leave every cache. By default, deletion removes the document's cached extractions (unless a surviving chunk has identical text), evicts its passages and all name vectors from the embedder's in-memory caches, drops query-trace lines that cite its chunks, and invalidates topic summaries that covered it. The optional model judge, which is off in every reported configuration, keeps a pair-decision cache that is not purged.

\subsection{Retrieval}
\label{sec:retrieval}
The graph has entity and passage nodes. Passage--entity edges come from forms, entity--entity edges are weighted by the number of facts linking the pair, and synonym edges are weighted by similarity.
Given a question $q$:
\begin{enumerate}
  \item Embed $q$ with bge-base-en-v1.5 \citep{xiao2023cpack}.
  \item Seed a Personalized PageRank walk. The subject and object entities of the two facts most similar to $q$ receive mass equal to the fact similarity divided by the number of passages mentioning the entity. Entities named in the question receive mass $2$ divided by their passage count: an entity is named when one of its aliases occurs verbatim in $q$, by greedy longest match within punctuation-delimited segments, and contains a capitalised or numeric word other than the question's first word. The 100 passages most similar to $q$ receive mass up to $0.05$ after min--max normalisation.
  \item Run the walk with damping $0.9$ and take the 30 highest-scoring passages.
  \item Score these passages with a local cross-encoder (bge-reranker-base) and fuse the cross-encoder rank with the PageRank rank by reciprocal rank fusion with $k=60$ \citep{cormack2009rrf}.
\end{enumerate}
The fused ranking is the system's retrieval output. The answer model receives the ten most similar facts with their evidence quotes and the top 10 passages in ranked order, and returns a short answer span as JSON.
No step before the answer call uses a language model. Matching question words against the alias table takes 0.03\,ms per question; the alias lookup is rebuilt from the store after every index or delete operation (0.6\,s on MuSiQue), so it never holds a deleted document's names.
\section{Experimental setup}

\paragraph{Datasets.}
We use MuSiQue \citep{trivedi2022musique}, 2WikiMultiHopQA \citep{ho2020twowiki} and HotpotQA \citep{yang2018hotpotqa}, following HippoRAG's construction: a seeded sample of development questions, with the corpus formed by all supporting and distractor paragraphs of the sampled questions.
MuSiQue has 1{,}000 questions over 11{,}402 passages, 2WikiMultiHopQA 300 over 2{,}131, and HotpotQA 300 over 2{,}957.
For 2WikiMultiHopQA, the one dataset on which \sys{} trailed in our first comparison, we also run the standard 1{,}000-question sample over 6{,}392 passages for \sys{}, HippoRAG~2 and dense retrieval; the 300-question sample is its first 300 questions.
Retrieval on MuSiQue and 2WikiMultiHopQA is reported on questions 501--1000, which no setting was tuned on, and on all 300 HotpotQA questions. Answers are generated for questions 501--800 of MuSiQue and of the 1{,}000-question 2WikiMultiHopQA sample, and for all 300 HotpotQA questions. LightRAG and fast-graphrag were run only on the 300-question 2WikiMultiHopQA sample, so they have no held-out 2WikiMultiHopQA results.

\paragraph{Baselines and parity.}
We compare against HippoRAG~2, LightRAG and fast-graphrag, widely used, maintained systems from three families that produce a ranked passage list, and against dense retrieval.
Every system uses Llama~3.3 70B for extraction, gpt-5.6-luna with reasoning disabled for answers, and bge-base-en-v1.5 for embeddings.
LightRAG runs in its default \texttt{mix} mode on PostgreSQL storage, because its default file storage rewrites every graph and vector file for each inserted document (\S\ref{sec:analysis}).

\paragraph{Reranker controls.}
\sys{} fuses a local cross-encoder into its ranking, which none of the baselines use by default. To separate the graph from the reranker, we give dense retrieval, HippoRAG~2 and LightRAG the same cross-encoder (bge-reranker-base) over their top 30 passages, fused with their own rank by the same reciprocal rank fusion \sys{} uses (\S\ref{sec:retrieval}). These controls are marked ``+RR''.

\paragraph{Answer stage.}
Each system's own answer pipeline gives the answer model different context: \sys{} adds fact sentences with evidence quotes, HippoRAG~2 uses a chain-of-thought template, LightRAG adds entity and relation descriptions.
To compare retrieval rather than context formatting, the main answer results use one identical answer stage for every system: its top 10 passages in rank order, labelled \texttt{[P1]}--\texttt{[P10]}, and one prompt asking for the shortest answer span and always a best guess.
The answer stage was run twice per system to measure run-to-run noise.

\paragraph{Tuning protocol.}
\sys's original retrieval settings were selected on MuSiQue questions 1--500. The question-entity seeding of \S\ref{sec:retrieval} was added for this version: its weight and the alternative of reranking candidate facts were selected on questions 1--500 of MuSiQue and of the 1{,}000-question 2WikiMultiHopQA sample, and the chosen setting was recorded before any held-out question was run. HotpotQA was never used for tuning, and the answer stage was not tuned. Rows labelled ``\sys{} (v1 seeding)'' are the published configuration without question-entity seeding.

\paragraph{Metrics and statistics.}
We report recall@$k$ over gold supporting passages, and exact match (EM) and token F1 over answer spans with standard normalisation.
Differences are paired by question with 95\% bootstrap intervals \citep{efron1994bootstrap} and two-sided bootstrap $p$-values.
We correct for multiple comparisons with the Holm procedure, separately over all retrieval comparisons and over all answer comparisons, and call a difference significant only when its Holm-adjusted $p$-value is below 0.05.
Index cost counts every model call and prices its tokens at one fixed list price per model: \$0.10/\$0.32 per million input/output tokens for Llama~3.3 70B and \$0.20/\$1.20 for gpt-5.6-luna. This keeps provider routing from distorting the comparison.

\paragraph{Deletion audit.}
After the main query pass, a seeded random 10\% of documents is deleted through each system's own API, and every question is re-run.
We then fingerprint the post-deletion store and compare it with a fresh index of the surviving documents built in a separate process from the cached extractions, for \sys{} and for HippoRAG~2.

\paragraph{Hardware.}
All local computation, including embeddings, reranking and graph walks, ran on one Apple M3 Pro laptop with 18~GB of memory. Model calls went through OpenRouter.

\section{Results}
\label{sec:results}

\begin{table}[t]
\centering
\caption{Retrieval recall@$k$ on questions no setting was tuned on: MuSiQue and 2WikiMultiHopQA (1{,}000-question sample) questions 501--1000, HotpotQA all 300. ``+RR'': the system's top 30 passages reranked by \sys's cross-encoder and fused with its own rank. ``v1 seeding'': \sys{} without question-entity seeding, as first published. fast-graphrag returns at most 8 passages; LightRAG and fast-graphrag were run only on the 300-question 2WikiMultiHopQA sample (``--''). Best value per column in bold.}
\label{tab:retrieval}
\small
\setlength{\tabcolsep}{4.5pt}
\begin{tabular}{lccccccccc}
\toprule
 & \multicolumn{3}{c}{MuSiQue} & \multicolumn{3}{c}{2WikiMultiHopQA} & \multicolumn{3}{c}{HotpotQA} \\
\cmidrule(lr){2-4}\cmidrule(lr){5-7}\cmidrule(lr){8-10}
System & R@2 & R@5 & R@10 & R@2 & R@5 & R@10 & R@2 & R@5 & R@10 \\
\midrule
\sys{}           & \textbf{.503} & \textbf{.660} & \textbf{.770} & .723 & \textbf{.907} & \textbf{.964} & .802 & .955 & \textbf{.988} \\
\sys{} (v1 seeding) & .469 & .640 & .758 & .691 & .856 & .935 & .795 & .943 & \textbf{.988} \\
HippoRAG 2       & .473 & .613 & .714 & \textbf{.752} & .903 & .930 & .777 & .930 & .975 \\
HippoRAG 2 +RR   & .460 & .611 & .713 & .678 & .794 & .899 & \textbf{.848} & \textbf{.957} & \textbf{.988} \\
LightRAG         & .313 & .477 & .584 & -- & -- & -- & .703 & .902 & .950 \\
LightRAG +RR     & .415 & .547 & .636 & -- & -- & -- & \textbf{.848} & .937 & .970 \\
fast-graphrag    & .310 & .426 & .475 & -- & -- & -- & .513 & .732 & .800 \\
Dense            & .398 & .545 & .625 & .644 & .714 & .758 & .768 & .908 & .955 \\
Dense +RR        & .444 & .567 & .660 & .666 & .730 & .772 & .838 & .942 & .968 \\
\bottomrule
\end{tabular}
\end{table}

\begin{table}[t]
\centering
\caption{Answer accuracy with an identical answer stage for every system (top 10 passages, same prompt and model) on held-out questions: MuSiQue and 2WikiMultiHopQA questions 501--800, HotpotQA all 300; first of two runs. Best value per column in bold. No difference between \sys{} and HippoRAG~2, with or without the reranker, is significant (Table~\ref{tab:paired}).}
\label{tab:answers}
\small
\begin{tabular}{lcccccc}
\toprule
 & \multicolumn{2}{c}{MuSiQue} & \multicolumn{2}{c}{2WikiMultiHopQA} & \multicolumn{2}{c}{HotpotQA} \\
\cmidrule(lr){2-3}\cmidrule(lr){4-5}\cmidrule(lr){6-7}
System & EM & F1 & EM & F1 & EM & F1 \\
\midrule
\sys{}         & \textbf{.383} & \textbf{.516} & .553 & .652 & .650 & \textbf{.804} \\
\sys{} (v1 seeding) & .363 & .505 & \textbf{.557} & \textbf{.654} & .643 & .795 \\
HippoRAG 2     & .363 & .504 & .550 & .643 & .630 & .776 \\
HippoRAG 2 +RR & .333 & .473 & .513 & .618 & \textbf{.653} & .800 \\
LightRAG       & .280 & .408 & -- & -- & .633 & .779 \\
Dense +RR      & .333 & .460 & .487 & .555 & .640 & .786 \\
\bottomrule
\end{tabular}
\end{table}

\begin{table}[t]
\centering
\caption{\sys{} minus HippoRAG~2 and minus HippoRAG~2 with \sys's reranker, paired by question, in percentage points with 95\% bootstrap intervals, on the held-out questions of Tables~\ref{tab:retrieval} and~\ref{tab:answers}. Bold: significant after Holm correction over all retrieval (36, including the comparisons with \sys{} v1 seeding and dense retrieval with the reranker) or all answer (28) comparisons.}
\label{tab:paired}
\small
\setlength{\tabcolsep}{4pt}
\begin{tabular}{lllll}
\toprule
Baseline & Metric & MuSiQue & 2WikiMultiHopQA & HotpotQA \\
\midrule
HippoRAG 2 & R@2  & $+3.0$ [$+0.4$, $+5.4$] & $-3.0$ [$-5.3$, $-0.8$] & $+2.5$ [$-1.2$, $+6.2$] \\
           & R@5  & $\mathbf{+4.8}$ [$+2.4$, $+7.1$] & $+0.4$ [$-1.6$, $+2.4$] & $+2.5$ [$+0.5$, $+4.5$] \\
           & R@10 & $\mathbf{+5.5}$ [$+3.5$, $+7.5$] & $\mathbf{+3.4}$ [$+1.9$, $+4.9$] & $+1.3$ [$+0.0$, $+2.7$] \\
           & EM   & $+2.0$ [$-2.0$, $+6.0$] & $+0.3$ [$-3.3$, $+4.0$] & $+2.0$ [$-1.3$, $+5.7$] \\
           & F1   & $+1.2$ [$-2.4$, $+4.9$] & $+0.9$ [$-2.6$, $+4.3$] & $+2.8$ [$-0.1$, $+5.6$] \\
\midrule
HippoRAG 2 +RR & R@2  & $\mathbf{+4.3}$ [$+2.3$, $+6.3$] & $\mathbf{+4.5}$ [$+2.9$, $+6.1$] & $-4.7$ [$-7.8$, $-1.5$] \\
               & R@5  & $\mathbf{+5.0}$ [$+3.0$, $+7.1$] & $\mathbf{+11.3}$ [$+9.5$, $+13.2$] & $-0.2$ [$-1.8$, $+1.5$] \\
               & R@10 & $\mathbf{+5.7}$ [$+3.7$, $+7.6$] & $\mathbf{+6.5}$ [$+5.0$, $+8.1$] & $+0.0$ [$-0.8$, $+1.0$] \\
               & EM   & $+5.0$ [$+1.3$, $+9.0$] & $+4.0$ [$+0.0$, $+8.0$] & $-0.3$ [$-3.0$, $+2.3$] \\
               & F1   & $+4.2$ [$+0.6$, $+7.9$] & $+3.4$ [$-0.2$, $+7.1$] & $+0.4$ [$-1.7$, $+2.5$] \\
\bottomrule
\end{tabular}
\end{table}

\paragraph{Retrieval.}
Table~\ref{tab:retrieval} reports recall and Table~\ref{tab:paired} the paired comparison with the strongest baseline.
After Holm correction, \sys{} is better than HippoRAG~2 on held-out MuSiQue at recall@5 (+4.8 points) and recall@10 (+5.5), and on held-out 2WikiMultiHopQA at recall@10 (+3.4). HippoRAG~2 keeps an edge at 2WikiMultiHopQA recall@2 (.752 against .723, $-3.0$ points, not significant after correction), and recall@5 is level there (.907 against .903). On HotpotQA the two are tied.
Question-entity seeding produces most of this change. Against the v1 configuration it adds 3.1, 5.1 and 2.9 points at recall@2, 5 and 10 on 2WikiMultiHopQA and 3.4 points at recall@2 on MuSiQue (all significant), and does not change HotpotQA. With v1 seeding, \sys{} trailed HippoRAG~2 significantly at 2WikiMultiHopQA recall@2 and recall@5.
\sys{} is therefore competitive with HippoRAG~2 on retrieval, ahead on MuSiQue and at the deeper ranks of 2WikiMultiHopQA, with no model call at query time. It exceeds LightRAG and fast-graphrag at recall@5 on every dataset on which they were run.

\paragraph{Is the gain the reranker's?}
The cross-encoder is the largest single component in our ablations (\S\ref{sec:analysis}), so we gave it to the baselines.
On the two datasets that need chains of evidence, the graph carries most of the gain: \sys{} exceeds dense retrieval with the same reranker by 9.3 points recall@5 on MuSiQue and 17.8 on 2WikiMultiHopQA (both significant after correction), and the reranker does not help HippoRAG~2 there (MuSiQue .613 to .611, 2WikiMultiHopQA .903 to .794).
The 2WikiMultiHopQA drop is not an artefact of the fusion constant. The cross-encoder alone ranks HippoRAG~2's top 30 at .736 recall@5, because it scores each passage against the question text and demotes second-hop passages that the question never names. Fusing with $k=10$ or $k=1$ instead of 60 gives .831 and .880, still below HippoRAG~2's own .903, so no fusion setting turns the reranker into a gain there. The +RR rows are therefore a control for the reranker's contribution, not an attempt to improve HippoRAG~2.
HotpotQA is different. Dense retrieval with the reranker reaches .942 recall@5, close to \sys's .955, and the reranked HippoRAG~2 is the best system at the top, 4.7 points above \sys{} at recall@2 (not significant after correction). On HotpotQA most supporting passages can be found from the question text, so the graph adds little and fusing the walk rank costs \sys{} some precision at the very top.

\begin{table}[t]
\centering
\caption{Recall under four embedders on held-out questions (MuSiQue and 2WikiMultiHopQA questions 501--1000, HotpotQA all 300). Extractions are fixed (each system reuses its bge-base extraction output), so only the embedder and the similarity thresholds change; \sys{} uses the same question-entity seeding under every embedder. Qwen3-Embedding-8B and text-embedding-3-large (``te-3-large'') are truncated to 1{,}024 dimensions. Dense +RR: dense top-30 with \sys's cross-encoder and fusion; for bge-large it uses the hosted model, whose vectors match the local model's (cosine 1.000 on every MuSiQue passage).}
\label{tab:embedders}
\small
\setlength{\tabcolsep}{3.5pt}
\resizebox{\linewidth}{!}{%
\begin{tabular}{llccccccccc}
\toprule
 & & \multicolumn{3}{c}{MuSiQue} & \multicolumn{3}{c}{2WikiMultiHopQA} & \multicolumn{3}{c}{HotpotQA} \\
\cmidrule(lr){3-5}\cmidrule(lr){6-8}\cmidrule(lr){9-11}
Embedder & System & R@2 & R@5 & R@10 & R@2 & R@5 & R@10 & R@2 & R@5 & R@10 \\
\midrule
bge-base & \sys & .503 & .660 & .770 & .723 & .907 & .964 & .802 & .955 & .988 \\
 & HippoRAG 2 & .473 & .613 & .714 & .752 & .903 & .930 & .777 & .930 & .975 \\
 & Dense +RR & .444 & .567 & .660 & .666 & .730 & .772 & .838 & .942 & .968 \\
\midrule
bge-large & \sys & .501 & .664 & .768 & .724 & .903 & .965 & .815 & .952 & .987 \\
 & HippoRAG 2 & .489 & .642 & .725 & .750 & .920 & .945 & .802 & .942 & .980 \\
 & Dense +RR & .431 & .582 & .672 & .671 & .741 & .777 & .845 & .942 & .970 \\
\midrule
Qwen3-Emb.-8B & \sys & .507 & .683 & .795 & .724 & .905 & .967 & .808 & .960 & .992 \\
 & HippoRAG 2 & .499 & .669 & .755 & .732 & .910 & .936 & .777 & .933 & .988 \\
 & Dense +RR & .469 & .637 & .737 & .674 & .753 & .799 & .863 & .947 & .967 \\
\midrule
te-3-large & \sys & .514 & .694 & .805 & .718 & .893 & .969 & .825 & .962 & .990 \\
 & HippoRAG 2 & .502 & .663 & .745 & .754 & .918 & .939 & .775 & .940 & .983 \\
 & Dense +RR & .443 & .628 & .721 & .677 & .763 & .815 & .838 & .942 & .968 \\
\bottomrule
\end{tabular}}
\end{table}

\paragraph{Does the comparison depend on the embedder?}
Every result above uses bge-base-en-v1.5, which is much smaller than the 7B embedder in the HippoRAG~2 paper. We re-indexed both systems with three larger embedders (Table~\ref{tab:embedders}): bge-large-en-v1.5, Qwen3-Embedding-8B and OpenAI text-embedding-3-large, the last two through OpenRouter with the provider pinned. Each system reused its own bge-base extraction output, so only the embedder changed.
Fixed similarity thresholds do not transfer between embedders, whose similarity scales differ. We therefore mapped each threshold without labels: 4.5\% of the 39{,}731 extracted surface names have a same-type nearest neighbour above 0.92 under bge-base and 27.2\% above 0.80, and for each new embedder we chose the thresholds that give the same shares (merge/synonym 0.912/0.809 for bge-large, 0.976/0.934 for Qwen3, 0.881/0.707 for text-embedding-3-large). HippoRAG~2's synonym threshold was mapped in the same way, and \sys's retrieval settings were not re-tuned.
The comparison keeps its shape under every embedder. With Holm correction over the 54 comparisons in the sweep, \sys{} leads HippoRAG~2 at MuSiQue recall@10 under every embedder (+4.0 to +6.1 points) and at 2WikiMultiHopQA recall@10 under Qwen3 and text-embedding-3-large (+3.1 and +3.0), all significant. HippoRAG~2 keeps an edge at 2WikiMultiHopQA recall@2 ($-0.9$ to $-3.6$ points, significant only with text-embedding-3-large), and recall@5 there is within 2.5 points either way. On HotpotQA \sys{} is ahead in all nine cells, never significantly.
Question-entity seeding helps under every embedder: on 2WikiMultiHopQA it adds 2.7 to 5.3 points of recall@5 over the v1 configuration with the same embedder, and on MuSiQue 2.0 to 3.5 points of recall@2.
Larger embedders help both systems on MuSiQue: recall@5 rises from .660 to .694 for \sys{} and from .613 to .669 for HippoRAG~2, so HippoRAG~2 gains more, and bge-large alone helps it but not \sys, plausibly because the cross-encoder dominates \sys's final ranking.
The graph's advantage over dense retrieval with the reranker persists: at recall@5 it is 13.0 to 16.2 points on 2WikiMultiHopQA under every embedder and 4.7 to 8.2 points on MuSiQue. On HotpotQA, Qwen3 dense retrieval with the reranker beats \sys{} by 5.5 points at recall@2.

\paragraph{Answers.}
With an identical answer stage on held-out questions (Table~\ref{tab:answers}), \sys{} and HippoRAG~2, with or without the reranker, tie on every dataset: the largest difference is 5.0 points EM, against the reranked HippoRAG~2 on MuSiQue, and none is significant. Better retrieval did not produce better answers: \sys{} and its v1 configuration differ by at most 2.0 points EM. \sys{} answers better than LightRAG on MuSiQue (+10.3 EM) and than dense retrieval with the reranker on 2WikiMultiHopQA (+9.7 F1), both significant.
In our first version each system used its own answer pipeline, and \sys{} appeared 7.7 points EM better on HotpotQA and 8.4 points F1 worse on 2WikiMultiHopQA. Both differences disappear under the identical stage, so they came from context formatting: \sys's fact sentences with evidence quotes on one side, HippoRAG~2's chain-of-thought template on the other.
Two identical runs of the answer stage changed the correctness of 9 to 22 of 300 answers per system and dataset, so answer differences of a few points on 300 questions are within run-to-run noise.

\begin{figure}[t]
\centering
\includegraphics[width=\linewidth]{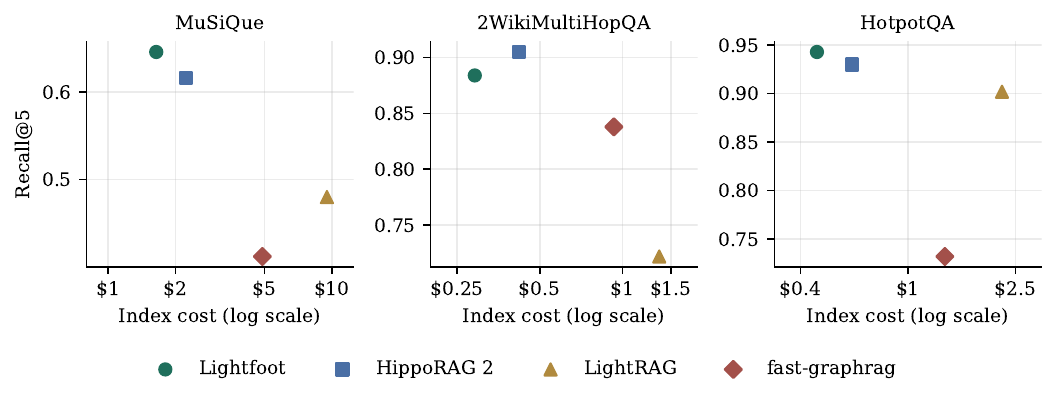}
\caption{Recall@5 against normalised index cost for each dataset (all MuSiQue questions; \sys{} with v1 seeding). Points nearer the top left are better.}
\label{fig:cost}
\end{figure}

\paragraph{Query path, latency and cost.}
Retrieval in \sys{} makes no model call. Each baseline makes at least one per question: keyword extraction (LightRAG), fact filtering (HippoRAG~2) or query entity extraction (fast-graphrag), so its retrieval latency and availability depend on the model provider.
Median retrieval-only latency on MuSiQue, excluding the answer call, was 0.53\,s for \sys, 0.81\,s for LightRAG, 1.56\,s for HippoRAG~2 and 1.96\,s for fast-graphrag.
Table~\ref{tab:cost} and Figure~\ref{fig:cost} show index cost. \sys{} has the lowest index cost of the four graph systems on every dataset, 26--31\% below HippoRAG~2, which makes two extraction calls per passage, and 3--6$\times$ below LightRAG and fast-graphrag. In absolute terms all of these are small at this scale; the difference that grows with traffic and with deletion volume is the number of model calls on the query and delete paths, which is zero for \sys.

\begin{table}[t]
\centering
\caption{Index cost at normalised list prices, and model calls per passage at index time and per question at retrieval time. LightRAG's MuSiQue cost is estimated from its 28{,}529 extraction and summary calls at measured average tokens per call, because its extraction was billed across several runs.}
\label{tab:cost}
\small
\begin{tabular}{lrrrrr}
\toprule
System & MuSiQue & 2WikiMultiHopQA & HotpotQA & Index calls & Retrieval calls \\
\midrule
\sys{}        & \textbf{\$1.64} & \textbf{\$0.29} & \textbf{\$0.46} & 1.0 & \textbf{0} \\
HippoRAG 2    & \$2.23 & \$0.42 & \$0.62 & 2.0 & 1 \\
fast-graphrag & \$4.89 & \$0.93 & \$1.37 & 1.1 & 1 \\
LightRAG      & $\approx$\$9.5 & \$1.36 & \$2.23 & 2.5 & 1 \\
\bottomrule
\end{tabular}
\end{table}

\begin{table}[t]
\centering
\caption{Deleting a seeded 10\% of MuSiQue documents (1{,}140 of 11{,}402). Recall@5 over all gold passages, on questions whose evidence survived (intact) and on questions with at least one deleted gold passage; the non-zero ``deleted, after'' values come from those questions' surviving gold passages, since no deleted passage was retrieved by any system. ``= rebuild'': the post-deletion store equals a fresh index of the survivors.}
\label{tab:delete}
\small
\setlength{\tabcolsep}{4pt}
\begin{tabular}{lrrcccc}
\toprule
System & Time & Model calls & Intact, before & Intact, after & Deleted, after & = rebuild \\
\midrule
\sys{}        & 14 s     & 0     & .648 & .661 & .369 & yes \\
HippoRAG 2    & 9 s      & 0     & .625 & .639 & .333 & yes$^\ast$ \\
LightRAG      & 13{,}294 s & 2{,}187 & .491 & .507 & .292 & not tested \\
fast-graphrag & \multicolumn{6}{c}{no deletion API; rebuilding from scratch is required} \\
\bottomrule
\multicolumn{7}{l}{\footnotesize $^\ast$On MuSiQue; not on a corpus of near-duplicate names (\S\ref{sec:results}, ``Where exactness breaks'').}
\end{tabular}
\end{table}

\paragraph{Deletion.}
Table~\ref{tab:delete} reports the deletion audit on MuSiQue, run with the v1 seeding configuration; question-entity seeding stores nothing, since its alias lookup is rebuilt from the store after each operation, so the exactness results below hold unchanged.
For every system that supports deletion, recall on questions whose evidence survived did not decrease, and none of the 229 deleted gold passages was retrieved afterwards by any system.
LightRAG re-summarises the descriptions of every entity touched by a deleted document, which took 2{,}187 model calls and 3.7 hours; on 2WikiMultiHopQA and HotpotQA it took 165 and 330 calls.
\sys{} and HippoRAG~2 made no model calls. \sys's deletion took 74\,s in our first version, which committed one transaction per document; deleting all documents in one transaction takes 14\,s and leaves an identical store.

\paragraph{Exactness at full scale.}
After the deletion, the \sys{} store was fingerprinted in one process and a fresh index of the 10{,}262 surviving documents was built in a second process from the extraction cache, with no model calls.
The fingerprints were identical, and every table matched exactly: 68{,}455 entities, 71{,}195 aliases, 106{,}472 mentions, 92{,}608 facts, 94{,}032 fact sources and 63{,}514 synonym edges.
We repeated the audit with the number guard (\S\ref{sec:analysis}) on the same 1{,}140 documents: the deletion took 15 seconds and the store again matched a fresh guarded index of the survivors in every table (68{,}705 entities, 67{,}154 synonym edges).
An earlier \sys{} version that resolved new names against the current canonical entities and kept top-$k$ synonym lists differed from the fresh index by 12 entities and 340 synonym edges.
We ran the same audit on HippoRAG~2. Its post-deletion graph and a fresh HippoRAG~2 index of the survivors have the same 80{,}738 nodes, the same chunk, entity and fact stores, and the same 265{,}641 edges with the same fact sources; 16 synonym weights differ by about $10^{-7}$, because its embeddings are computed in batches whose composition differs between the two builds. On MuSiQue, HippoRAG~2's deletion is therefore exact up to floating-point noise.

\paragraph{Where exactness breaks.}
HippoRAG~2 links each entity to at most 100 nearest neighbours above a similarity threshold. On none of the three benchmark corpora does the cap bind: the largest number of synonym neighbours is 55 on MuSiQue, 50 on 2WikiMultiHopQA and 18 on HotpotQA. Corpora with many near-identical names, such as customer accounts, product codes or ticket identifiers, do bind it.
We built a 160-passage corpus in which each passage names one customer account and one order, with fixed extractions given identically to both systems, deleted every fourth document, and compared with a fresh index of the 120 survivors.
HippoRAG~2's post-deletion graph had 7{,}558 synonym edges against 8{,}123 in the rebuild: survivors lost the edges to deleted neighbours and never gained the next-nearest survivors, which a fresh build links. \sys{}, whose synonym edges are a pure threshold, matched its rebuild in every table.
The same corpus exposes a weakness in \sys's entity resolution. Without the number guard, the 0.92 links chain account names into components: 201 of the 240 names fell into 73 entities that each joined different accounts or orders, the largest joining 27 accounts. Deletion was exact, but the entities were wrong. With the guard, every name is its own entity and the deletion audit again matches the rebuild in every table. The cost is edge growth: without a cap, synonym edges among mutually similar names grow quadratically, from 1{,}661 to 8{,}128 here, which is the growth HippoRAG~2's cap bounds. A cap that stays exact would have to be a function of each pair alone, such as a stricter threshold within dense clusters, which we did not test.
The difference between the two systems is thus not whether deletion calls a model, but whether every derived structure is free of history; top-$k$ lists are an easily missed violation.

\section{Analysis}
\label{sec:analysis}

\paragraph{What changes accuracy.}
Table~\ref{tab:levers} collects every change we measured, in one place, with the size of its effect and where it was measured. Positive values are improvements. Effects on the tuning halves are reported only as evidence for the design; the held-out results in \S\ref{sec:results} are the test.

\begin{table}[t]
\centering
\caption{What changes accuracy in \sys, from the experiments in this paper. R@$k$ in points of recall@$k$; ``tune'': questions 1--500; ``held-out'': questions 501--1000 (MuSiQue, 2WikiMultiHopQA) and all of HotpotQA. ``n.s.'': not significant after Holm correction.}
\label{tab:levers}
\small
\resizebox{\linewidth}{!}{%
\begin{tabular}{>{\raggedright\arraybackslash}p{55mm}>{\raggedright\arraybackslash}p{58mm}>{\raggedright\arraybackslash}p{30mm}l}
\toprule
Change & Effect & Measured on & Status \\
\midrule
\multicolumn{4}{l}{\emph{Raises retrieval}} \\
Graph walk vs.\ dense retrieval, same reranker & R@5 $+9.3$ MuSiQue, $+17.8$ 2Wiki; $+1.3$ HotpotQA (n.s.) & held-out & core design \\
Question-entity seeding (weight 2) & 2Wiki R@2/5/10 $+3.1$/$+5.1$/$+2.9$; MuSiQue R@2 $+3.4$; HotpotQA 0 & held-out & on \\
Fuse cross-encoder with walk rank (vs.\ replace) & R@5 $+3.4$ & MuSiQue tune & on \\
Damping 0.5 $\to$ 0.9 (with fusion) & R@5 $+2$ to $+3$ & MuSiQue tune & on \\
Seed from 2 facts instead of 5 & R@5 about $+1$ & MuSiQue tune & on \\
\midrule
\multicolumn{4}{l}{\emph{No reliable effect on retrieval}} \\
Number guard on entity links & at most $\pm0.1$ at every $k$; fixes wrong merges & held-out & option \\
Fact nodes $\to$ weighted entity edges & 0; graph 40\% smaller & MuSiQue tune & on \\
Model judge for names at cosine 0.80--0.92 & R@5 within 1 (noise); 5{,}626 fewer calls when removed & MuSiQue 2k slice & removed \\
Larger embedders (bge-large, Qwen3-8B, te-3-large) & comparison with HippoRAG~2 keeps its shape & held-out & -- \\
\midrule
\multicolumn{4}{l}{\emph{Lowers retrieval}} \\
Rerank candidate facts before seeding (with question entities) & R@2 $-1.5$ to $-1.7$ (tune), $-1.2$ to $-1.6$ (held-out) & MuSiQue, 2Wiki & off \\
Fusing the walk rank on HotpotQA & reranked HippoRAG~2 $4.7$ above \sys{} at R@2 (n.s.) & HotpotQA & -- \\
\midrule
\multicolumn{4}{l}{\emph{Answers (EM/F1)}} \\
Identical answer stage for all systems & removed v1's $+7.7$ EM (HotpotQA) and $-8.4$ F1 (2Wiki) gaps & held-out & used \\
Top 10 ranked passages vs.\ 8 + entity notes & HotpotQA EM $+2.3$ [$+0.3$, $+4.7$]; others within noise & own pipeline & on \\
Hybrid BM25 + dense channel & gold in context $+2.1$; no reliable answer gain; 35\% more tokens & MuSiQue slice & off \\
Retrieval gains above & no significant answer difference vs.\ HippoRAG~2 & held-out & -- \\
Re-running the same answer stage & changes 9--22 of 300 answers (noise floor) & held-out & -- \\
\bottomrule
\end{tabular}}
\end{table}

\paragraph{Ablations.}
All ablations were run on MuSiQue questions 1--500 without model calls.
Fusing the cross-encoder rank with the PageRank rank, instead of replacing the PageRank order, was the largest single gain at 3.4 points of recall@5.
Raising damping from 0.5 to 0.9 under fusion added 2 to 3 points, and seeding from 2 facts rather than 5 added about 1.
Removing fact nodes from the graph, by turning facts into weighted entity--entity edges, changed nothing and shrank the graph by 40\%.
On a 2{,}000-passage slice, removing the model-based judge for name pairs with similarity 0.80 to 0.92 changed recall@5 by less than one point, within noise across 141 questions. It saved 5{,}626 judge calls on the full corpus.
The reranker controls of \S\ref{sec:results} complete this picture: fusion is the largest gain within \sys, but the same reranker without the graph falls 9 to 18 points short on the multi-hop datasets.

\paragraph{Entity resolution and the number guard.}
The chaining seen on the synthetic corpus also occurs on the benchmark corpora, mostly among dates and counts. Without the guard, 211 MuSiQue entities join names with different numbers (560 names; the largest joins 26 dates such as ``10 July'' and ``14 June''), as do 64 on 2WikiMultiHopQA (``47 films'' with ``48 films'') and 24 on HotpotQA. With the guard these fall to 4, 2 and 1; the remainder are chained through a name without digits, such as ``20 episodes'' to ``twenty episodes'' to ``26 episodes''.
We rebuilt the three stores with the guard from the extraction cache and the stored vectors, so that nothing else changed; a control rebuild without the guard reproduced the published HotpotQA store in every table. Held-out recall changed by at most 0.1 points on every dataset and $k$ (for example MuSiQue .503/.661/.770 against .503/.660/.770), so the tables, computed without the guard, stand. The released code enables the guard as an option; turning it on changes the index and requires a rebuild.

\paragraph{Seed selection.}
For this version we tested two model-free replacements for HippoRAG~2's model-based fact filter, on questions 1--500 of MuSiQue and of the 2WikiMultiHopQA sample. Seeding entities named in the question raised recall@2 by 3.9 points on 2WikiMultiHopQA and 2.6 on MuSiQue at weight 2; the gain flattened between weights 1 and 4. Reranking the 20 most similar facts with the cross-encoder before linking the top 2 to 5 raised 2WikiMultiHopQA recall@10 by up to 3.7 points but left MuSiQue unchanged, and combined with question-entity seeding it lowered recall@2 on both datasets by 1.5 to 1.7 points, so it is off. On the held-out questions the combination again cost 1.2 to 1.6 points at recall@2.

\paragraph{Answer context.}
In \sys's own answer pipeline, replacing eight passages plus entity descriptions with the top 10 passages in ranked order raised HotpotQA EM by 2.3 points, with a 95\% interval of $+0.3$ to $+4.7$. Changes on the other two datasets were within noise.
Adding a separate hybrid channel, Elasticsearch BM25 and dense search fused by reciprocal rank fusion, raised the share of gold passages in the context from 0.852 to 0.873 on a MuSiQue slice. It added 35\% to answer tokens and produced no reliable answer gain.
Every way of selecting 10 passages from the union of the graph and hybrid rankings covered fewer gold passages than the graph ranking alone.

\paragraph{Where \sys{} fails.}
On the 300 answered MuSiQue questions, in the run that also included the hybrid channel, 183 answers were wrong. In 109 of these at least one gold passage was missing from the context, and 74 had every gold passage present.
Missing evidence concentrates on later hops. The entity of the first hop's passage is named in the question 83\% of the time, against 36\% for the second hop and 28\% for the third and fourth.
Search over the question text, lexical or dense, has little to match these passages against; following the chain is the more likely route to them.
Seeding from the two most similar facts sometimes selects incidental facts such as release dates over the bridging fact. HippoRAG~2's model-based fact filter addresses this, which plausibly explains the advantage it had at recall@2 and recall@5 on 2WikiMultiHopQA, where most questions compare two named entities; seeding those named entities directly closes most of that gap.
Of the 74 failures with complete evidence, 49 were near misses with F1 of at least 0.5: over-long spans, answers naming an intermediate entity, and off-by-one dates.

\paragraph{Baseline failure modes.}
Two silent failures would have corrupted the comparison had they gone unnoticed.
LightRAG's first index aborted most of its insertion pipeline after an embedding timeout but returned normally, leaving 31\% of chunks indexed. Its first score (recall@5 0.239) was discarded, and the reported run verifies that every document reached the processed state.
LightRAG's default file storage rewrote all graph and vector files for each inserted document, about 415\,MB per document at 3{,}000 passages, so throughput fell continuously.
HippoRAG~2 aborted its entire index when 24 of 11{,}402 extraction outputs were malformed JSON. A tolerant parser in our adapter recovered 11{,}397; five passages were indexed without entities.

\section{Limitations}
\begin{itemize}
  \item \sys's retrieval settings were tuned on the first halves of MuSiQue and the 1{,}000-question 2WikiMultiHopQA sample, so retrieval and answers are reported only on the second halves and on HotpotQA. The deletion audit and the 300-question 2WikiMultiHopQA comparison with LightRAG and fast-graphrag use the v1 seeding configuration.
  \item Answer metrics rest on 300 questions per dataset, and HotpotQA uses a 300-question sample; run-to-run noise changes the correctness of 3--7\% of answers.
  \item All systems share one extraction model. The main comparison uses one small embedder, so HippoRAG~2's absolute scores are below those published with its 7B embedder; the embedder sweep (Table~\ref{tab:embedders}) reuses extractions, maps thresholds by a label-free calibration, and covers retrieval only. We did not test whether the cost ordering holds for other extraction models.
  \item On HotpotQA the graph adds no retrieval benefit over dense retrieval with a reranker.
  \item Question-entity seeding was designed after v1 trailed HippoRAG~2 on 2WikiMultiHopQA. Its weight was chosen on the tuning halves and its test was pre-registered, but the choice of what to build was informed by benchmark results. The number guard was added after a reviewer pointed out single-linkage chaining; it was chosen from the synthetic corpus, not tuned on any benchmark.
  \item The case in which HippoRAG~2's deletion diverges is a synthetic 160-passage corpus; the cap binds on none of the three benchmark corpora. It isolates a mechanism we expect in identifier-heavy data, but we had no real enterprise corpus, did not measure how often the cap binds in real catalogues, and did not fingerprint LightRAG's deletion. All corpora have at most 11{,}402 passages and were indexed on one laptop.
  \item Without a cap, synonym edges grow quadratically within clusters of mutually similar names. On the benchmark corpora there are at most about one synonym edge per entity (1.03 on MuSiQue), but a catalogue of millions of near-identical identifiers would need a pairwise rule to bound it.
  \item Corpus-level summarisation questions, the target of community-summary systems, were not evaluated.
\end{itemize}

\section{Conclusion}
\sys{} shows that a GraphRAG index can need no model at retrieval time, support exact, model-free deletion, and be cheap to build, while remaining competitive with HippoRAG~2, the strongest baseline in our comparison, under identical models: better on MuSiQue retrieval and at the deeper ranks of 2WikiMultiHopQA, tied on HotpotQA, and tied on answer accuracy everywhere. Seeding entities named in the question, by exact alias match, closes most of the gap to HippoRAG~2 on 2WikiMultiHopQA comparison questions without its model-based fact filter.
Exact deletion is not automatic for systems that already delete without model calls: it requires every derived structure, including neighbour lists, to be free of history.
The remaining retrieval gap lies at the very top of the ranking on 2WikiMultiHopQA and in reaching later hops that the question never names; better retrieval has not yet translated into better answers, so the practical case for \sys{} rests on query cost, index cost and deletion. Model-free second-hop retrieval from entities found in the first round is the most direct next step.

\paragraph{Reproducibility.}
Code, the benchmark harness, baseline adapters, run configurations, the question and passage subsets, and per-question predictions and cost ledgers for every reported run are available at \url{https://github.com/udayallu-ml/Lightfoot-GraphRAG}.

\paragraph{Use of AI tools.}
An AI assistant was used to implement parts of the benchmarking harness, to run experiments and to draft this manuscript. The authors reviewed all results and text and take full responsibility for them.

\bibliographystyle{plainnat}
\bibliography{references}

\appendix
\section{Survey of open-source GraphRAG frameworks}
\label{app:survey}
Table~\ref{tab:survey} summarises the source-code review of 17 open-source frameworks that motivates this work (reviews in the repository under \texttt{docs/frameworks}). Families: A community summaries, B keyword dual-level retrieval, C PageRank over extracted structure, D agentic or planner-based, E graph-database SDKs, T summary tree. Call counts are model calls in the default configuration. ``Document delete'' means an API that removes one source document and what was derived from it; no framework documents or tests that the result equals an index that never contained the document.

\begin{table}[h]
\centering
\caption{Index and query model calls and lifecycle support in 17 open-source GraphRAG frameworks.}
\label{tab:survey}
\scriptsize
\resizebox{\linewidth}{!}{%
\begin{tabular}{llllll}
\toprule
Framework & Fam. & Index calls / chunk & Query calls & Incremental insert & Document delete \\
\midrule
Microsoft GraphRAG & A & $\sim$2 + 1 per community & 1 (local), dozens (global) & yes, no re-clustering & no \\
nano-graphrag & A & $\sim$2 + reports per insert & 1 to many & full re-cluster & no \\
R2R & A & 0.5 + 1 per community & 1 & communities rebuilt & partial, stale copies \\
LightRAG & B & 2 & 2 & yes & yes, model re-summary \\
PathRAG & B & 2 & 2 & hash dedup only & entity only \\
MiniRAG & B & 2, whole corpus per insert & 2 & impractical & entity only \\
HyperGraphRAG & B & 3--5 & 2 & append only & entity only, broken \\
RAGFlow & B & $\sim$2 per batch & 0--1 & manual sync & yes \\
HippoRAG 2 & C & 2 & 1--2 & yes & yes, provenance \\
fast-graphrag & C & 1 & 2 & yes & no (upsert policies) \\
Graph-R1 & D & 3--5 & 2--3 & none & no \\
KAG & D & 3--4 & 6--15 & yes & yes \\
Youtu-GraphRAG & D & 1 + 1 per 5 communities & 2--8 & none & no \\
GraphRAG-SDK & E & $\sim$1 & 2 & yes, crash-safe & yes \\
neo4j-graphrag & E & 1 & 1 & none & no \\
TrustGraph & E & $\sim$2 & 2 & append only & collection only \\
RAPTOR & T & 0.1--0.5 & 1 & none & no \\
\bottomrule
\end{tabular}}
\end{table}

\end{document}